\documentclass{svproc}
\usepackage{url}
 \usepackage[hidelinks]{hyperref}
\usepackage{graphicx}%
\usepackage{multirow}%
\usepackage{amsmath,amssymb,amsfonts}%
\usepackage{mathrsfs}%
\usepackage[title]{appendix}%
\usepackage{xcolor}%
\usepackage{textcomp}%
\usepackage{manyfoot}%
\usepackage{booktabs}%
\usepackage{listings}%
\usepackage{arydshln}
\usepackage{graphics}
\usepackage{amssymb}
\usepackage{amsfonts}
\usepackage[linesnumbered,lined,commentsnumbered]{algorithm2e}
\usepackage{algpseudocode}
\usepackage{textcomp}
\usepackage{xcolor}
\usepackage{adjustbox}
\usepackage{placeins}
\usepackage{booktabs} 
\usepackage{lipsum}
\usepackage{float}
\usepackage{multirow}
\usepackage[normalem]{ulem}
\usepackage[caption=false]{subfig}
\usepackage{epsfig}
\usepackage{amsmath}
\usepackage{mathtools}
\usepackage{longtable}

\begin{document}
\mainmatter              
\title{A Conditioned Unsupervised Regression Framework Attuned to the Dynamic Nature of Data Streams}
\titlerunning{Unsupervised Regression Framework}  
%


\author{Ren\'e Richard\inst{1} \and Nabil Belacel\inst{2}}
\authorrunning{Richard et al.} 
%
\tocauthor{René Richard, Nabil Belacel}
\institute{
National Research Council, Fredericton, New Brunswick, Canada,\\
\email{rene.richard@nrc-cnrc.gc.ca}
\and
National Research Council, Ottawa, Ontario, Canada,\\
\email{nabil.belacel@nrc-cnrc.gc.ca}
}

\maketitle              

\begin{abstract}

In scenarios where obtaining real-time labels proves challenging, conventional approaches may result in sub-optimal performance. This paper presents an optimal strategy for streaming contexts with limited labeled data, introducing an adaptive technique for unsupervised regression. The proposed method leverages a sparse set of initial labels and introduces an innovative drift detection mechanism to enable dynamic model adaptations in response to evolving patterns in the data. To enhance adaptability, we integrate the ADWIN (ADaptive WINdowing) algorithm with error generalization based on Root Mean Square Error (RMSE). ADWIN facilitates real-time drift detection, while RMSE provides a robust measure of model prediction accuracy. This combination enables our multivariate method to effectively navigate the challenges of streaming data, continuously adapting to changing patterns while maintaining a high level of predictive precision. We evaluate the performance of our multivariate method across various public datasets, comparing it to non-adapting baselines. Through comprehensive assessments, we demonstrate the superior efficacy of our adaptive regression technique for tasks where obtaining labels in real-time is a significant challenge. The results underscore the method's capacity to outperform traditional approaches and highlight its potential in scenarios characterized by label scarcity and evolving data patterns.

\end{abstract}

\section{Introduction}
\label{sec:introduction}

The goal in a regression problem is to model the relationship between one or more independent variables (also known as the input variables) and a dependent variable (also known as the target variable). A model is trained on input features, where the objective is to minimize the squared distance of the target variable when compared to the predicted output. In order to determine the squared distance, the value for the target variable must be known. In this context, the target variable is commonly referred to as the label or ground truth label, which is a continuous quantity. The training scenario outlined here pertains to a supervised linear regression problem.

While some applications may have an abundance of labeled training data, others, in principle, possess a vast training set, yet the size of the labeled subset is relatively small. The challenge of acquiring class labels may stem from either limited knowledge or restricted resources, as deriving a target variable for corresponding input features often necessitates expensive expertise.  In addition, obtaining labels in real-time poses a significant challenge or may be deemed impractical. When feasible, obtaining genuine labels is subject to delay. In industrial practice for example, acquiring a single authentic label can take at least a few hours~\cite{ZHANG20231}. In contrast, acquiring unlabeled data is more straightforward as it demands less effort, expertise, and time.

Machine learning models discern patterns from historical data. These discerned patterns, often called 'concepts,' play a vital role in the identification of relationships among input and output variables. When variable relationships change, the patterns models have learned may become invalid. The term concept drift, or dataset shift, is sometimes used as a generic term to describe any changes in the statistical properties found in data. In the literature, authors use different names to refer to the same notion, or use the same name for different notions \cite{baena2006early}. Moreno-Torres et al.~\cite{moreno2012unifying} argue that inconsistent terminology is a disservice to the field (e.g. makes literature searches difficult and confounds discussions on the topic). For the purposes of this work, we  apply the definition of concept drift and causes found in~\cite{ZHANG20231}. 

According to~\cite{ZHANG20231}, concept drift is triggered by differences in distribution between the historical training data and online deployment data and can produce two kinds of process drifts : concept drift and virtual drift. The concept drift implies the relationship between input and output variables has changed over time, potentially triggered by shifts in the operational conditions of, for instance, an industrial process. The virtual drift implies the distribution of the input variables has changed over time, which may be caused by a degradation of, for instance,  sensors in the aforementioned industrial process. By exploring various sources and types of drift, we gain insights into the underlying causes of the more severe instances, enabling the diagnosis of potential model issues and facilitating the selection of the most effective path for swift resolution. 

Industry 4.0 and the Industrial Internet of Things (IIoT) rely on automation, machine learning (ML), and real-time data. Models are trained offline and deployed for near real-time prediction, but evolving input-output relationships necessitate periodic retraining and redeployment. With the rise of streaming IIoT process data, there's a growing demand for near real-time analytics. Adaptive techniques address issues like concept drift, providing early insights and simplifying production deployments. However, in real-world scenarios, the misconception of static environments for ML models can lead to performance issues due to dynamic data distributions.

Many regression models designed for streaming data predominantly operate under the test-then-train paradigm, assuming the timely availability of label data following the arrival of new data in the stream~\cite{vzliobaite2015optimizing}. While this approach is prevalent in the literature, it tends to be impractical in real-world scenarios. Despite the acknowledged disparity with reality in various practical problems, adaptive regression models persist in presuming that the label or ground truth becomes promptly accessible immediately after their predictions. This assumption, though convenient for model development, may lead to sub-optimal performance in situations where obtaining labels in general, let alone, in real-time is challenging or impractical. The optimal approach, in a streaming context with label scarcity, involves using an adaptive technique enabling unsupervised regression that leverages a small amount of labels initially, complemented by a drift detection method, enabling model adaptations to pattern changes in data.

To tackle the challenge of data distributions evolving over time, this work introduces an innovative algorithm that capitalizes on the capabilities of two distinct regression models trained on independent datasets of different sizes. Our approach is attuned to the dynamic nature of data streams and the potential for concept drift. Within our framework, we initially train one regression model, 
on an initial dataset of size $W$, and another model on a different independent dataset of size $n$. These two models collectively yield a linear representation of the target variable. By employing two models trained on different data sources, we tap into the diversity of information and enhance adaptability to changing data patterns. This diversity, in turn, forms a robust mechanism for estimating the truth value of the target variable when applied to incoming data. As new data flows into our streaming algorithm, we vigilantly monitor for drift. Upon detecting drift, the model is promptly retrained to align with the evolving data distribution, ensuring that it stays accurate and aligned with the changing reality of the data. Our approach embraces the concept of ensemble learning, where multiple models complement each other to provide more accurate and adaptable predictions. By amalgamating the strengths of different models trained on independent datasets and incorporating advanced drift detection, our algorithm offers a powerful solution for near real-time, adaptive, and unsupervised regression tasks. 

As a final introductory remark, it should be noted that the scientific contributions in this work are described as follows:

\begin{itemize}
    \item A novel unsupervised regression approach for streaming data using simple, explainable models.
    \item A novel drift detection method. 
    \item A comparative analysis of the proposed method relative to a non-adapting baseline.
\end{itemize}

The rest of the paper is organized as follows. Section~\ref{sec:related_work} provides a summary of related works. Section~\ref{sec:mat_and_method} describes the materials and method, the problem setting, the novel online anomaly detection approach proposed in this work in addition to the datasets used for benchmarking. Then in Section~\ref{sec:results}, we discuss the results. Finally, Section~\ref{sec:conclusion} concludes and indicates future research work.

\section{Related Works}
\label{sec:related_work}

Over time, the research community and industry have advanced data acquisition and sensor technologies to measure diverse physical properties and phenomena. As data sources become more accessible and the throughput of data transmission increases, there's a rising demand for real-time execution of traditional analytical tasks on raw data streams.

Gemaque et al.~\cite{gemaque2020overview} present a summary of unsupervised approaches used to identify concept drift in classification problems. The authors highlight that unsupervised approaches represent an under-explored research topic. A popular drift detection algorithm is the Drift Detection Method (DDM)~\cite{gama2004learning}. This method attempts to leverage the online error rate of the algorithm. Bifet et al. propose a new approach for dealing with distribution change and concept drift when learning from data sequences that may vary with time named ADWIN (short for  ADaptive WINdowing)~\cite{bifet2007learning}. The authors use sliding window sizes that adapt according to the rate of change observed in the data. ADWIN does not require the user to define as many specifications on the window sizes to observe changes. Many techniques or algorithms for adapting to concept drift in the literature either stem from or integrate with ADWIN~\cite{lu2018learning}.

Numerous established machine learning methodologies embrace the concept of online learning, a paradigm where models are continuously updated in real-time without reliance on ground truth labels. Online learning finds wide application in the realm of machine learning and data analysis, especially in scenarios involving streaming data and dynamic environments~\cite{novac2020toward}\cite{ravaglia2021tinyml}. While these approaches typically commence with an initial training phase, utilizing ground truth values for model calibration, they seamlessly transition into an online learning mode. In this mode, models evolve and adapt to the ever-changing data streams, even in the absence of ground truth labels. It's worth noting that despite the prevalence of online learning, only a limited subset of these techniques is dedicated to the domain of adaptive regression. The majority of existing approaches tend to be tailored for unsupervised online anomaly detection~\cite{belacelRichardXu2022lstm}\cite{munir2019fusead}\cite{vazquez2023anomaly}. 

Despite the focus on anomaly detection problems, recent advancements in the field of machine learning have seen the emergence of a limited number of unsupervised online regression techniques. Notably, Andrade et al.~\cite{andrade2023online} have introduced an approach aligned with the principles of the Tiny Machine Learning (TinyML) paradigm, emphasizing computational efficiency, low energy consumption, and suitability for resource-constrained devices. Their method involves an unsupervised, online linear regression algorithm, capable of continual learning as new data arrives, thus demonstrating adaptability to changing data patterns and concept evolution.  The regression algorithm introduced in this study, referred to as TEDA Regressor, is a modified version of the TEDA algorithm~\cite{angelov2014anomaly}, initially created for identifying data outliers. Additionally, it incorporates elements from the AutoCloud algorithm~\cite{bezerra2020evolving}, which is typically employed for data classification and clustering problems. Utilizing an unsupervised approach, the TEDA Regressor method generates dataclouds, and for each cloud, it utilizes an adaptive Recursive Least Squares (RLS) filter to predict the subsequent value in the time series data by considering past inputs. The adapted algorithm is utilized in the regression problem of predicting the real-time fuel consumption of vehicles. The authors characterize the proposed technique as a time series prediction method that fits within the concept of regression. The method does not focus on multi-variate input variables.

Another notable contribution is found in the work by Zhang et al.~\cite{ZHANG20231}, which delves into the realm of domain adaptation as an innovative mechanism for unsupervised knowledge calibration. This study presents a Gaussian mixture continuously adaptive regression approach designed for soft sensor modeling, with a primary focus on addressing concept drift adaptation issues. In contrast, our work stands out for its distinct emphasis on an online, adaptive, and unsupervised regression framework tailored for streaming data, addressing both virtual and real drift detection. While existing efforts concentrate on concept drift adaptation problems and soft sensor modeling, our focus encompasses a broader spectrum of streaming data scenarios.

\section{Materials and Method}
\label{sec:mat_and_method}

The proposed method in this work, described in Figure~\ref{fig:algo_window_overview}, draws inspiration from the insights presented in reference~\cite{DetectingVirtualConceptDriftRegression2021}, where the authors demonstrated the utility of employing various regression models to estimate the ground truth error for an ordinary least square linear regression model. In numerous real-world scenarios, the task of estimating an unknown target variable using regression models is commonplace.

\subsection{Background and problem setting}

Consider a streaming data scenario where the initial labeled data stream is denoted by $(x_t, y_t)$ for $t =1,\ldots, W$, where $x_t$ represents the feature vector at time $t$ and $y_t$ represents the corresponding label or the target variable. After the initial labeled window, only features $x_t$ are available without corresponding labels $y_t$ for $t> W$.
The objective is to develop an algorithm that adapts to evolving data distributions in real-time, ensuring accurate predictions of $y_t$ despite the challenges of a dynamic streaming environment where only ${x_t, t> W}$ are available. 
\subsubsection{Ordinary least squares for linear regression}
Data arrives in a sequence over time, predictions need to be made in real or near real-time. For the prediction we used linear regression~\cite{james_linear_2021} as presented in Equation \ref{eq:LR}. The most common approach to estimate the linear regression parameters for batch or streaming data is the Ordinary Least Squares (OLS)~\cite{zdaniuk_ordinary_2014}. In linear regression, we have a set of independent variables $x=(x_1, x_2,\ldots, x_n)$ and a dependent variable $y$ the relationship between the $n$ input variables $x_1, \ldots x_n$ and the target or dependent variable $y$ is linear and modeled as: 

\begin{equation}
\label{eq:LR}
y = \beta_0 + {\beta}_1x_1 + {\beta}_2x_2+\ldots+{\beta}_nx_n + \epsilon 
\end{equation}

where  
$y$ is the dependent variable;
$\beta_0$ is the intercept term;
$\beta_1, \beta_2,\ldots, \beta_n$ are the coefficients of the independent variables;
$x_1, x_2,\ldots, x_n$ are the independent variables;
$\epsilon$ is the error term.
The objective of OLS is to find the values of $\beta_0, \beta_1, \ldots, \beta_n$ that minimize the sum of squared differences between the observed and predicted values of $y$ as presented in Equation \ref{eq:OF}. 
\begin{equation}
\label{eq:OF}
    Minimize  \sum^{N}_{i=1}{\left ( y_i-(\beta_0+\beta_1x_{i1}+\ldots +\beta_nx_{in})\right )}
\end{equation}
where $N$ is the number of observations, $x_{ij}: i=1,\ldots N; j=1,\ldots n$ is the feature $j$ of sample $i$. 
Then OLS estimates the coefficients $\beta_0,\beta_1,\ldots,\beta_n$ by solving the normal equation:
\begin{equation}
\label{eq:esteq}
    X^TX\hat{\beta} = X^TY
\end{equation}
where $X$ is the matrix of independent variables; $Y$ is the vector of dependent variable values and $\hat{\beta}$ is the vector of estimated coefficients.
Hence, the solution is given by:
\begin{equation}
\label{eq:soleq}
    \hat{\beta} = (X^TX)^{-1}X^TY
\end{equation}
where $\hat{\beta_0}$ is the estimated intercept; $\hat{\beta_1}, \ldots, \hat{\beta_n}$ are the estimated coefficients for the independent variables.
Once the coefficients are estimated using OLS estimation with $n$ training set, predictions of dependent variable $\hat{y}$ for new data $x_{new}$ can be made using the linear equation:
 \begin{equation}
 \hat{Y}= \hat{\beta_0} + \hat{\beta_1}x_1 + \ldots +\hat{\beta_n}x_n\\
 \end{equation}
 
\begin{equation}
\label{eq:OLS}
 \hat{Y} = x_{new}\hat{beta}
 \end{equation}

We have chosen OLS to fit our models because it is straightforward and easy to interpret. It can be adapted to streaming data by employing an incremental learning approach. As new data points arrive, the model can be updated to reflect the latest information without requiring a complete retraining of the model. Furthermore, OLS maintains a relatively low memory footprint compared to more complex models. This can be advantageous in scenarios where memory resources are limited, common in streaming data environments. Scanning an entire dataset multiple times to update a model in a streaming scenario is impractical due to the large volume of data. To avoid multiple scans, a data stream is processed using windows. In a streaming data scenario, the concept of a sliding window is employed to focus on a subset of the most recent data while allowing for a dynamic and adaptable model. The sliding window contains the most recent $W$ data points, and as new data arrives, the window slides to incorporate the latest observations while removing the oldest ones.

\subsubsection{Handling concept drift with ADWIN and the generalization error of regression models}
In the ever-evolving landscape of streaming data, maintaining the accuracy and relevance of predictive models is paramount. To navigate this dynamic environment, we propose an advanced approach to online adaptive unsupervised regression. This approach leverages the principles of real and virtual concept drift, integrating active drift detection with ADWIN and absolute error RMSE for effective model adaptation in real-time.
The amalgamation of ADWIN for active drift detection and RMSE for absolute error analysis provides a robust foundation for understanding and responding to the intricacies of changing data dynamics. ADWIN actively monitors the streaming data, adapting its window size dynamically to detect both gradual and abrupt shifts in statistical properties. Simultaneously, RMSE quantifies the absolute error between predicted and actual values, offering insights into the model's predictive accuracy.
Our hybrid strategy automates the detection of drift events, triggering the seamless training of new models. By combining the strengths of ADWIN and RMSE, our model avoids overfitting to short-term fluctuations, ensuring adaptability to genuine shifts in the data distribution while maintaining resilience against noise and temporary patterns.
This comprehensive approach allows for timely model adaptation, where upon detecting either type of drift, the model is promptly updated. This ensures the ongoing accuracy and relevance of the regression model to the evolving data characteristics.
Implemented within the context of a sliding window, this approach strikes a harmonious balance between adaptability to recent data and robustness against short-term fluctuations. The goal is to enhance the accuracy and reliability of the regression model in streaming data scenarios, showcasing the efficacy of our methodology in the absence of ground truth values.

\begin{figure}[ht]
    \centering
    \includegraphics[width=0.9\linewidth]{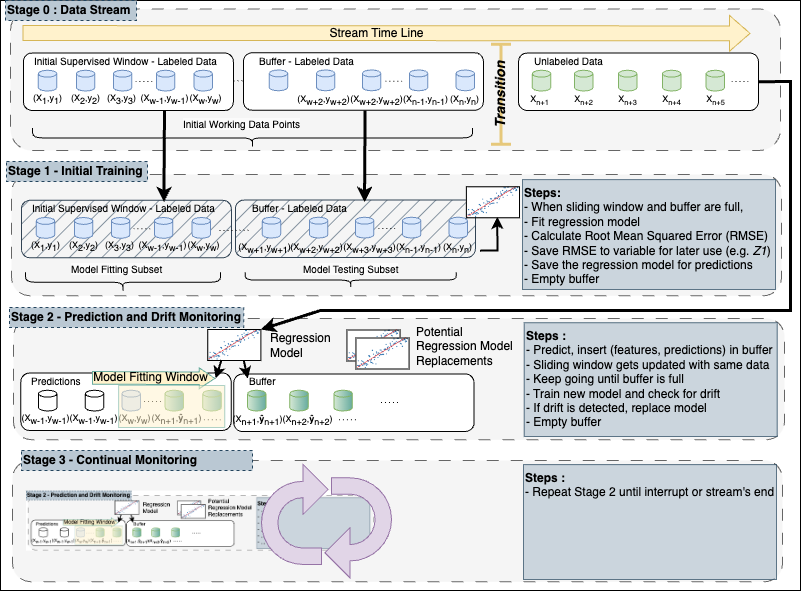}
    \caption{Online Adaptive Unsupervised Regression Framework}
    \label{fig:algo_window_overview}
\end{figure}

\subsection{Online adaptive unsupervised regression framework}
\label{sec:adaptive_regresion_model}

We propose an online, adaptive, and unsupervised regression framework for handling streaming data. Initially, we assume a supervised period with available target variable or ground truth values, followed by a transition to an unsupervised period where ground truth values are unavailable. In managing data streams, a typical strategy involves starting with labeled data and then adapting to newly arriving data. The full source code for our method is available on GitHub~\cite{misc_github_regression}. The general framework of our proposed algorithm is presented in Figure~\ref{fig:algo_window_overview} with the following steps:

\begin{itemize}

    \item \textit{Stage 0 - Data Stream}: The framework starts with acquiring a data stream. The stream is composed of two separate segments. One segment contains an initial set of working data points of length $n$, which contains labels (i.e. ground truth values). The second stream segment contains unlabeled samples.
    
    \item \textit{Stage 1 - Initial Training}: Split the initial working data points segment into two subsets. One subset will be used to fill a sliding window for model fitting. We refer to this window as the model fitting subset. The other will be used for testing and to calculate the Root Mean Squared Error (RMSE) as calculated in Equation~\ref{eq:rmse_eq}. We refer to this subset as the model testing subset. This subset is also referred to as the buffer in our method. As new labeled data arrives from the initial working data points, fill the model fitting  window, once this window is full, fill the model testing window (a.k.a. the buffer). Once the model fitting window and the buffer are full, fit a regression model (e.g. linear regression). During this preliminary training stage, the initial RMSE is calculated using the model's predictions on the testing data and the actual target values. This RMSE value is then saved. We refer to this value as the $Z1$ value in our method. Finally, the regression model is saved for predictions or replacement in the next stages and the buffer is cleared.

    \item \textit{Stage 2 - Predictions and Drift Monitoring}: Upon the arrival of a fresh, unlabeled sample, predict the target variable. Combine the input features of the unlabeled sample with the predicted target variable and add this to the buffer. Additionally, this same information is used to update the model fitting window by incorporating the new data into the sliding window and eliminating the oldest data from it. Once the buffer is full, fit the model using the model fitting window. Predict the target values for the data in the buffer using the new model. Save the newly trained model. We refer to this model as $M''$ in our method. Calculate a new RMSE metric by comparing the predicted values with the previous true values in the buffer. We refer to this new RMSE value as the $Z2$ value. Calculate the difference between the new RMSE and the previous RMSE (i.e. difference between $Z1$ and $Z2$). If the difference exceeds a predefined threshold, consider this a drift detection event, indicating that the model might need replacing due to a significant change in the data distribution. If a drift is detected, replace the current regression model with the new updated $M''$ regression model. Empty the buffer regardless of whether a drift is detected or not.  

    \item \textit{Stage 3 - Continual Monitoring} : As new data from the stream becomes available, repeat Stage 2. Continue monitoring for drift and update the model, the model fitting window and buffer as necessary.
    
\end{itemize}

To allow more sophisticated drift detection and model adaptation in our streaming regression, we have optionally incorporated the ADWIN detector in stage 2 of our proposed method. With this option, when ADWIN detects drift, the approach checks for further evidence of drift using RMSE and take appropriate action. When ADWIN doesn't detect drift, the algorithm continues without taking any action, emptying the buffer as needed.

\subsection{Datasets for benchmarking}
\label{sec:datasets}

We consider 3 regression datasets collected from the UC Irvine (UCI) Machine Learning Repository. The target variable in these regression datasets are used as ground truth information to evaluate the performance of the proposed method. We chose datasets of different sizes and subject areas to cover a range of applications and feature spaces. The individual characteristics of each dataset and experiment is summarized in Table~\ref{tab:datasets}.

\begin{table}[ht]
\begin{adjustbox}{max width=0.92\textwidth}
\begin{tabular*}{\textwidth}{@{\extracolsep\fill}llllll}
\toprule%
\toprule
Dataset \& Experiment & \vspace{2mm} & Features & Instances & Target Variable & Subject Area \\
\toprule
Air Quality - (Exp. 1) & \vspace{2mm} & 8 & 9,357 &  Carbon Monoxide (CO) & Comput. Sci. \\
\cmidrule(rl){1-6} 
Air Quality - (Exp. 2) & \vspace{2mm} & 8 & 9,357 & Nitrogen Dioxide (NO2) & Comput. Sci. \\
\cmidrule(rl){1-6}
\begin{tabular}{@{}l@{}} Air Quality - (Exp. 3) \\ \hspace{2mm} \end{tabular} & \vspace{2mm} & \begin{tabular}{@{}l@{}} 8 \\ \hspace{2mm} \end{tabular}  &\begin{tabular}{@{}l@{}} 9,357 \\ \hspace{2mm} \end{tabular}  & \begin{tabular}{@{}l@{}} Non Metanic \\ 
Hydrocarbons (NMHC)\end{tabular} & \begin{tabular}{@{}l@{}} Comput. Sci. \\ \hspace{2mm} \end{tabular} \\
\cmidrule(rl){1-6}
\begin{tabular}{@{}l@{}} Concrete - (Exp. 4) \\ \hspace{2mm} \end{tabular} & \vspace{2mm} & \begin{tabular}{@{}l@{}} 8 \\ \hspace{2mm} \end{tabular}  &\begin{tabular}{@{}l@{}} 1,030 \\ \hspace{2mm} \end{tabular}  & \begin{tabular}{@{}l@{}} Compressive \\ Strength \end{tabular} & \begin{tabular}{@{}l@{}} Physics \& \\ Chem. \\ \hspace{2mm} \end{tabular} \\
\cmidrule(rl){1-6}
Protein - (Exp. 5) & \vspace{2mm} & 9 & 45,730 & RMSD & Biology \\
\cmidrule(rl){1-6}
Turbine - (Exp. 6) & \vspace{2mm} & 11 & 36,734 & Turbine Energy Yield (TEY) & Comput. Sci. \\
\cmidrule(rl){1-6}
Turbine - (Exp. 7) & \vspace{2mm} & 10 & 36,734 & Carbon Monoxide (CO) & Comput. Sci. \\
\cmidrule(rl){1-6}
Turbine - (Exp. 8) & \vspace{2mm} & 10 & 36,734 & NOx (i.e. NO + NO2) &  Comput. Sci. \\
\bottomrule
\bottomrule
\end{tabular*}
\end{adjustbox}
\vspace{1mm}
\caption{Dataset Summary}
\label{tab:datasets}
\end{table}

\subsubsection{Air quality}
Experiments 1, 2, and 3 utilize the UCI air quality dataset~\cite{misc_air_quality_360}, comprised of 9,357 instances of hourly averaged readings obtained from an array of five metal oxide chemical sensors integrated into an air quality chemical multisensor device. The temporal range for the dataset is from March 2004 to February 2005 and represents the longest freely available recordings of on-field deployed air quality chemical sensor device responses. Hourly averaged concentrations ground truth values of Carbon monoxide (CO), Non Metanic Hydrocarbons (NMHC), Benzene, Total Nitrogen Oxides (NOx), and Nitrogen Dioxide (NO2) were obtained from a reference certified analyzer co-located with the study. With the availability of multiple ground truth values, three separate regression experiments were conducted with this dataset. As described in Tables~\ref{tab:datasets},~\ref{tab:experiment_parameters} and~\ref{tab:lr_results} the target variables in these experiments are Carbon Monoxide (CO), Nitrogen Dioxide (NO2) and Non Metanic Hydrocarbons (NMHC).

\subsubsection{Concrete}
Experiment 4 utilizes the UCI concrete dataset~\cite{misc_concrete_compressive_strength_165}, comprised of 1,030  instances with nine numerical features and one target variable. The target variable is the compressive strength of the concrete. The numerical features are composed of the concrete's ingredients and age.

\subsubsection{Protein}
Experiment 5 uses the UCI protein dataset~\cite{misc_physicochemical_properties_of_protein_tertiary_structure_265}, which contains the physicochemical properties of protein tertiary structures. The dataset is comprised of 45,730 decoys and sizes vary from 0 to 21 armstrong. The target variable is the Root Mean Square Deviation (RMSD), which is a similarity indicator in protein structure prediction algorithms.

\subsubsection{Turbine}
Finally, experiments 6, 7, and 8 utilize the UCI turbine dataset~\cite{misc_gas_turbine_co_and_nox_emission_data_set_551}, comprised of 36,734 instances of 11 sensor measurements aggregated over one hour, from a gas turbine located in Turkey. This dataset includes gas turbine parameters such as turbine inlet temperature and compressor discharge pressure in addition to  ambient variables. Three separate regression experiments were conducted with this dataset. Each individual experiment, in turn, predicting a target variable of either Turbine Energy Yield (TEY), Carbon Monoxide (CO) or Oxides of Nitrogen (NOx), which include Nitric Oxide (NO) and Nitrogen Dioxide (NO2).

\newpage


\begin{table}[H]
\begin{adjustbox}{max width=0.92\textwidth}
\begin{tabular}{lllllllll}
\toprule 
\midrule

\multicolumn{1}{c}{} & \multicolumn{4}{c}{\textbf{AIR QUALITY}} & \multicolumn{4}{c}{\textbf{AIR QUALITY}} \\ 
\multicolumn{1}{c}{} & \multicolumn{4}{c}{Target Variable : CO (GT) } & \multicolumn{4}{c}{Target Variable : NO2 (GT)} \\
\cmidrule{3-5} \cmidrule{7-9} 
Parameter &  & Exp. 1 - (a) & Exp. 1 - (b) & Exp. 1 - (c) &  & Exp. 2 - (a) & Exp. 2 - (b) & Exp. 2 - (c)  \\
\cmidrule{1-1} \cmidrule{3-5} \cmidrule{7-9} 
Working Data Points &  & 120 & 120 & 120 &  & 120 & 120 & 120   \\
\cdashline{3-5} \cdashline{7-9} \\
\begin{tabular}{lll}Secondary Drift  \\ Detector \\ Configuration \end{tabular}   &  & \begin{tabular}{lll}N/A \\ \hspace{1mm}  \\ \hspace{1mm}  \end{tabular} & \begin{tabular}{lll}ADWIN\hspace{3mm} \\ Features: 1 \\ Delta: 0.1e-15\end{tabular} & \begin{tabular}{lll}N/A \\ \hspace{1mm}  \\ \hspace{1mm}  \end{tabular} &  & \begin{tabular}{lll}N/A \\ \hspace{1mm}  \\ \hspace{1mm}  \end{tabular} & \begin{tabular}{lll}ADWIN\hspace{3mm} \\ Features: 1 \\ Delta: 0.1e-15\end{tabular} & \begin{tabular}{lll}N/A \\ \hspace{1mm}  \\ \hspace{1mm}  \end{tabular} \\
\cdashline{3-5} \cdashline{7-9}
\\
Is Baseline  &  & False & False & True &  & False & False & True \\
\cdashline{3-5} \cdashline{7-9}
\\
RMSE Delta Threshold  &  & 0.1e-4 & 0.1e-4 & N/A &  & 0.1e-4 & 0.1e-4 & N/A   \\
\cdashline{3-5} \cdashline{7-9}
\\
\begin{tabular}{lll}Regression  \\ Algorithm \\ \hspace{1mm} \end{tabular}  &  & \begin{tabular}{lll}Scikit-learn \\ Linear \\ Regression\end{tabular} & \begin{tabular}{lll}Scikit-learn \\ Linear \\ Regression\end{tabular} & \begin{tabular}{lll}Scikit-learn \\ Linear \\ Regression\end{tabular} &  & \begin{tabular}{lll}Scikit-learn \\ Linear \\ Regression\end{tabular} & \begin{tabular}{lll}Scikit-learn \\ Linear \\ Regression\end{tabular} & \begin{tabular}{lll}Scikit-learn \\ Linear \\ Regression\end{tabular} \\

\midrule
\\

\multicolumn{1}{c}{} & \multicolumn{4}{c}{\textbf{AIR QUALITY}} & \multicolumn{4}{c}{\textbf{CONCRETE}} \\ 
\multicolumn{1}{c}{} & \multicolumn{4}{c}{Target Variable : NMHC (GT) } & \multicolumn{4}{c}{Target Variable : Compressive Strength} \\
\cmidrule{3-5} \cmidrule{7-9} 
Parameter &  & Exp. 3 - (a) & Exp. 3 - (b) & Exp. 3 - (c) &  & Exp. 4 - (a) & Exp. 4 - (b) & Exp. 4 - (c)  \\
\cmidrule{1-1} \cmidrule{3-5} \cmidrule{7-9} 
Working Data Points &  & 120 & 120 & 120 &  & 120 & 120 & 120   \\
\cdashline{3-5} \cdashline{7-9} \\
\begin{tabular}{lll}Secondary Drift  \\ Detector \\ Configuration \end{tabular}   &  & \begin{tabular}{lll}N/A \\ \hspace{1mm}  \\ \hspace{1mm}  \end{tabular} & \begin{tabular}{lll}ADWIN\hspace{3mm} \\ Features: 1 \\ Delta: 0.1e-15\end{tabular} & \begin{tabular}{lll}N/A \\ \hspace{1mm}  \\ \hspace{1mm}  \end{tabular} &  & \begin{tabular}{lll}N/A \\ \hspace{1mm}  \\ \hspace{1mm}  \end{tabular} & \begin{tabular}{lll}ADWIN\hspace{3mm} \\ Features: 1 \\ Delta: 0.1e-15\end{tabular} & \begin{tabular}{lll}N/A \\ \hspace{1mm}  \\ \hspace{1mm}  \end{tabular} \\
\cdashline{3-5} \cdashline{7-9}
\\
Is Baseline  &  & False & False & True &  & False & False & True \\
\cdashline{3-5} \cdashline{7-9}
\\
RMSE Delta Threshold  &  & 0.1e-4 & 0.1e-4 & N/A &  & 0.3 & 0.3 & N/A   \\
\cdashline{3-5} \cdashline{7-9}
\\
\begin{tabular}{lll}Regression  \\ Algorithm \\ \hspace{1mm} \end{tabular}  &  & \begin{tabular}{lll}Scikit-learn \\ Linear \\ Regression\end{tabular} & \begin{tabular}{lll}Scikit-learn \\ Linear \\ Regression\end{tabular} & \begin{tabular}{lll}Scikit-learn \\ Linear \\ Regression\end{tabular} &  & \begin{tabular}{lll}Scikit-learn \\ Linear \\ Regression\end{tabular} & \begin{tabular}{lll}Scikit-learn \\ Linear \\ Regression\end{tabular} & \begin{tabular}{lll}Scikit-learn \\ Linear \\ Regression\end{tabular} \\

\midrule
\\

\multicolumn{1}{l}{} & \multicolumn{4}{c}{\textbf{PROTEIN}} & \multicolumn{4}{c}{\textbf{TURBINE}} \\ 
\multicolumn{1}{l}{} & \multicolumn{4}{c}{Target Variable : RMSD } & \multicolumn{4}{c}{Target Variable : TEY} \\ 
\cmidrule{3-5} \cmidrule{7-9} 
Parameter &  & Exp. 5 - (a) & Exp. 5 - (b) & Exp. 5 - (c) &  & Exp. 6 - (a) & Exp. 6 - (b) & Exp. 6 - (c)  \\
\cmidrule{1-1} \cmidrule{3-5} \cmidrule{7-9} 
Working Data Points &  & 120 & 120 & 120 &  & 120 & 120 & 120   \\
\cdashline{3-5} \cdashline{7-9} \\

\begin{tabular}{lll}Secondary Drift  \\ Detector \\ Configuration \end{tabular}   &  & \begin{tabular}{lll}N/A \\ \hspace{1mm}  \\ \hspace{1mm}  \end{tabular} & \begin{tabular}{lll}ADWIN\hspace{3mm} \\ Features: 1 \\ Delta: 0.1e-16\end{tabular} & \begin{tabular}{lll}N/A \\ \hspace{1mm}  \\ \hspace{1mm}  \end{tabular} &  & \begin{tabular}{lll}N/A \\ \hspace{1mm}  \\ \hspace{1mm}  \end{tabular} & \begin{tabular}{lll}ADWIN\hspace{3mm} \\ Features: 1 \\ Delta: 0.1e-15\end{tabular} & \begin{tabular}{lll}N/A \\ \hspace{1mm}  \\ \hspace{1mm}  \end{tabular} \\

\cdashline{3-5} \cdashline{7-9} \\
Is Baseline  &  & False & False & True &  & False & False & True \\
\cdashline{3-5} \cdashline{7-9} \\
RMSE Delta Threshold  &  & 0.15 & 1.2e-14 & N/A &  & 0.1e-11 & 0.1e-15 & N/A   \\
\cdashline{3-5} \cdashline{7-9} \\
\begin{tabular}{lll}Regression  \\ Algorithm \\ \hspace{1mm} \end{tabular}  &  & \begin{tabular}{lll}Scikit-learn \\ Linear \\ Regression\end{tabular} & \begin{tabular}{lll}Scikit-learn \\ Linear \\ Regression\end{tabular} & \begin{tabular}{lll}Scikit-learn \\ Linear \\ Regression\end{tabular} &  & \begin{tabular}{lll}Scikit-learn \\ Linear \\ Regression\end{tabular} & \begin{tabular}{lll}Scikit-learn \\ Linear \\ Regression\end{tabular} & \begin{tabular}{lll}Scikit-learn \\ Linear \\ Regression\end{tabular} \\

\midrule
\\

\multicolumn{1}{l}{} & \multicolumn{4}{l}{\textbf{TURBINE}} & \multicolumn{4}{l}{\textbf{TURBINE - (Exp. 8)}} \\
\multicolumn{1}{l}{} & \multicolumn{4}{l}{Target Variable : CO } & \multicolumn{4}{l}{Target Variable : NOX} \\ 
\cmidrule{3-5} \cmidrule{7-9} 
Parameter &  & Exp. 7 - (a) & Exp. 7 - (b) & Exp. 7 - (c) &  & Exp. 8 - (a) & Exp. 8 - (b) & Exp. 8 - (c)  \\
\cmidrule{1-1} \cmidrule{3-5} \cmidrule{7-9} 
Working Data Points &  & 120 & 120 & 120 &  & 120 & 120 & 120  \\
\cdashline{3-5} \cdashline{7-9} \\

\begin{tabular}{lll}Secondary Drift  \\ Detector \\ Configuration \end{tabular}   &  & \begin{tabular}{lll}N/A \\ \hspace{1mm}  \\ \hspace{1mm}  \end{tabular} & \begin{tabular}{lll}ADWIN\hspace{3mm} \\ Features: 1 \\ Delta: 0.1e-15\end{tabular} & \begin{tabular}{lll}N/A \\ \hspace{1mm}  \\ \hspace{1mm}  \end{tabular} &  & \begin{tabular}{lll}N/A \\ \hspace{1mm}  \\ \hspace{1mm}  \end{tabular} & \begin{tabular}{lll}ADWIN\hspace{3mm} \\ Features: 1 \\ Delta: 0.1e-15\end{tabular} & \begin{tabular}{lll}N/A \\ \hspace{1mm}  \\ \hspace{1mm}  \end{tabular} \\
\cdashline{3-5} \cdashline{7-9} \\
Is Baseline  &  & False & False & True &  & False & False & True \\
\cdashline{3-5} \cdashline{7-9} \\
RMSE Delta Threshold  &  & 0.1e-15 & 0.1e-15 & N/A &  & 0.1e-15 & 0.1e-15 & N/A   \\
\cdashline{3-5} \cdashline{7-9} \\
\begin{tabular}{lll}Regression  \\ Algorithm \\ \hspace{1mm} \end{tabular}  &  & \begin{tabular}{lll}Scikit-learn \\ Linear \\ Regression\end{tabular} & \begin{tabular}{lll}Scikit-learn \\ Linear \\ Regression\end{tabular} & \begin{tabular}{lll}Scikit-learn \\ Linear \\ Regression\end{tabular} &  & \begin{tabular}{lll}Scikit-learn \\ Linear \\ Regression\end{tabular} & \begin{tabular}{lll}Scikit-learn \\ Linear \\ Regression\end{tabular} & \begin{tabular}{lll}Scikit-learn \\ Linear \\ Regression\end{tabular} \\

\bottomrule
\bottomrule
\end{tabular}
\end{adjustbox}
\vspace{1mm}
\caption{Parameter Summary of the 24 Experiments}
\label{tab:experiment_parameters}
\end{table}


\subsection{Experiments and computational environment}
\label{sec:experiments}

The experiments were conducted on a Mac M2 Pro, equipped with a 12-core CPU, a 19-core GPU, and 32GB of RAM, ensuring a robust computational environment. The framework and experiments in this research are written in Python, and leverage open-source libraries such as Scikit-learn~\cite{pedregosa2011scikit}, Pandas~\cite{mckinney2010data}, Matplotlib~\cite{hunter2007matplotlib}, and River~\cite{montiel2021river}. Docker~\cite{boettiger2015introduction} was utilized to package the software into a standardized assembly referred to as a container. The source code and containerized experimental environment used this work can be accessed via GitHub~\cite{misc_github_regression}.

To ensure the reproducibility of our experimental results, the process to launch the experiments and collect the results was containerized and automated via parameterized scripts. In total, 24 experiments were run. These are detailed in Table~\ref{tab:experiment_parameters}. In the table, the section headings outline the experiment's input data such as the \textit{AIR QUALITY, CONCRETE, PROTEIN} datasets, etc. Additionally, the predicted target variable is also described in each section heading. The experimental parameters are listed on the left-hand side of the table. These parameters include settings such as the number of  \textit{Working Data Points, the Secondary Drift Detector Configuration}, etc. The systematic permutations of these parameters, in combination with the input data and target variables, yield 24 experiments, which are listed as \textit{Exp. 1 - (a)}, \textit{Exp. 1 - (b)} \textit{Exp. 1 - (c)}, etc. in the table.

The online, adaptive regression framework expects a minimum number of labeled data points to prime the predictive algorithm. We refer to these as working data points in this work. After these initial data points are processed, the framework no longer needs any labeled data to make predictions and adjust for perceived drifts in the input data.

\section{Discussion of the Results}
\label{sec:results}

In this section, experimental results found in Table~\ref{tab:lr_results} are discussed. Table~\ref{tab:lr_results} follows the same overall format found in Table~\ref{tab:experiment_parameters}. To get a feel for the general performance of each model, when applied to a specific dataset, we used the popular Root Mean Squared Error (RMSE) metrics found in regression problem evaluations. We briefly define the evaluation metrics found in Table~\ref{tab:lr_results} next. For a visual overview of selected results, a scatter plot for some experiments, depicting predicted vs. ground truth values and model retraining events, can also be found in appendix ~\ref{ap:1} to ~\ref{ap:4}.

\subsection{Drift detector}
\label{sec:drift_detector_results}

This value is really a modeling parameter. It was included in the results table to improve the readability of the results. This parameter has three options : \textit{RMSE Absolute Error, ADWIN} and \textit{RMSE Absolute Error} and \textit{None}. The first option allows the model to adjust to evolving data stream patterns using custom, windowed, intermediary, RMSE evaluations. The second option, combines the first option with the ADWIN drift detector implementation from the River Python library. Finally, the third option essentially specifies that no model updates can occur and can be seen as a baseline model option. In summary, this parameter affects whether the model can be updated or not based on drifts detected in the data stream.

\subsection{Model updates}
\label{sec:model_updates_results}

This metric describes how often a drift in the data stream was detected, triggering a model retraining and update. In this work, the drift detection algorithm uses predictions to determine drifts in the data streams. No ground truth information is assumed in this process. Retraining a model takes time but may increase predictive performance. Decisions regarding the trade-off between execution time and predictive performance should be timed optimally within the data stream.

\subsection{Root mean squared error}
\label{sec:results_rmse}
The purpose of the RMSE metric in regression problems is to quantify how well a model's predictions align with the actual observed values. It does so by calculating the average error magnitude between predicted and actual values, giving more weight to larger errors through the square root operation. RMSE values are expressed in the same units as the target variable, making it easy to interpret. Lower RMSE values indicate better model performance. RMSE values are often used during optimization processes in order to obtain better predictive modeling results. The RMSE metric formula is defined in equation~\ref{eq:rmse_eq}. It is important to note that RMSE calculations rely on available ground truth information, which we possess for reporting purposes, and is evaluated as a comprehensive assessment at the conclusion of processing the entire data stream.

\begin{equation} 
RMSE(Y, y, B) = \sqrt{\sum_{i=1}^{n}\left( Y_i-y_i\right)^{2}}
\label{eq:rmse_eq}
\end{equation}

\subsection{Total execution time}
\label{sec:results_execution_time}

This represents the total execution time (in seconds) for processing an entire data stream's instances. The lower the reported values here, the better.

\subsection{Processing Rate}
\label{sec:results_processing_rate}
This represents the per-data point time (in records per seconds) taken to processing a stream's instance. The lower the reported values here, the better.

\subsection{Results}
In this section, we discuss in detail the experimental results found in Table~\ref{tab:lr_results}. We describe the best results in order of the input dataset and according to the best RMSE performance metric values. 

\subsubsection{Air quality}
For air quality-related results, experiments \textit{1-(a), 2-(a) and 3-(a)} produced the best results. The \textit{RMSE Absolute Error} framework method produced the closest air quality target variable predictions for CO, NO2, NHMC as indicated by the RMSE metric values highlighted in bold. The selective retraining based on intermediary, windowed RMSE comparisons (i.e. RMSE Absolute Error) produced the highest number of model update events for air quality-related experiments. Specifically, for the CO target variable predictions, the model was updated 14 times. For the NO2 and NMHC predictions, the model was updated 18 and 20 times respectively. The RMSE value differences between the best predictive method and the alternatives ranges from 0.04 to 3 units for these experiments. 

\subsubsection{Concrete}
For concrete-related results, experiment \textit{4-(a)} produced the best result. Again, the \textit{RMSE Absolute Error} framework method produced the closest overall concrete compressive strength target variable predictions as indicated by the RMSE metric value highlighted in bold. The model was only updated 3 times during the course of processing this stream. The RMSE value differences between the best predictive method and the alternatives is around 2.9 units for the concrete-related experiments.

\subsubsection{Protein}
Experiment \textit{5-(a)} produced the best result for protein-related experiments. The \textit{RMSE Absolute Error} framework method produced the closest Root Mean Square Deviation (RMSD) target variable predictions as indicated by the RMSE metric value (highlighed in bold). In addition, the model was only updated 3 times during the course of processing this stream. Finally, the RMSE value differences between the best predictive method and the alternatives is around 0.2 units for the protein-related experiments.

\subsubsection{Turbine}
For turbine-related results, experiments \textit{6-(a), 7-(b)(c) and 8-(a)} produced the best results. The best Turbine TEY target variable predictions were produced with the \textit{RMSE Absolute Error} framework method. The number of model retraining events for this dataset and method was 38. The best Turbine CO target variable predictions were produced by both the \textit{ADWIN \& RMSE Absolute Error} and the baseline where the drift detection is \textit{None} methods. These methods produced the similar RMSE metric values. In experiment 7, a number of retraining events did not impact the performance over the baseline where no drift detection and model retraining was performed. Finally, the best regression results for predicting Turbine NOX were produced with the \textit{RMSE Absolute Error} framework method. 

\newpage

\begin{table}[H]
\centering
\begin{adjustbox}{max width=0.92\textwidth}
\begin{tabular}{lllllllll}
\toprule 
\midrule

\multicolumn{1}{c}{} & \multicolumn{4}{c}{\textbf{AIR QUALITY}} & \multicolumn{4}{c}{\textbf{AIR QUALITY}} \\ 
\multicolumn{1}{c}{} & \multicolumn{4}{c}{Target Variable : CO (GT) } & \multicolumn{4}{c}{Target Variable : NO2 (GT)} \\
\multicolumn{1}{l}{} & \multicolumn{4}{c}{Target Variable Range : [0.1 : 11.9] } & \multicolumn{4}{c}{Target Variable Range : [2.0 : 340.0]} \\ 
\multicolumn{1}{l}{} & \multicolumn{4}{c}{ Prediction Count : 9,237 } & \multicolumn{4}{c}{Prediction Count : 9,237} \\ 
\cmidrule{3-5} \cmidrule{7-9} 
Metric &  & Exp. 1 - (a) & Exp. 1 - (b) & Exp. 1 - (c) &  & Exp. 2 - (a) & Exp. 2 - (b) & Exp. 2 - (c)  \\
\cmidrule{1-1} \cmidrule{3-5} \cmidrule{7-9} 
\begin{tabular}{lll}Drift  \\ Detector\\ \hspace{1mm} \end{tabular}  &  & \begin{tabular}{lll}RMSE \\ Abs. \\ Error\end{tabular} & \begin{tabular}{lll}ADWIN \& \\ RMSE Abs.\\ Error\end{tabular} & \begin{tabular}{lll}None \\ \hspace{1mm} \\ \hspace{1mm}\end{tabular} &  & \begin{tabular}{lll}RMSE \\ Abs. \\ Error\end{tabular} & \begin{tabular}{lll}ADWIN \& \\ RMSE Abs.\\ Error\end{tabular} & \begin{tabular}{lll}None \\ \hspace{1mm} \\ \hspace{1mm}\end{tabular} \\
\cdashline{3-5} \cdashline{7-9} \\
Model Updates &  & 14 & 1 & 0 &  & 18 & 1 & 0  \\
\cdashline{3-5} \cdashline{7-9}
\\
RMSE &  & \textbf{1.4125\textsuperscript{*}} & 1.4582 & 1.4582 &  & \textbf{35.1157\textsuperscript{*}} & 38.1399 & 38.1399  \\
\cdashline{3-5} \cdashline{7-9}
\\
Total Execution Time (Seconds) &  & 3.09 & 2.00 & 0.92 &  & 3.15 & 2.03 & 0.91  \\
\cdashline{3-5} \cdashline{7-9}
\\
Processing Rate (Records/Second) &  & 2,989 & 4,618 & 10,040 &  & 2,932 & 4,550 & 10,150 \\

\midrule
\\

\multicolumn{1}{l}{} & \multicolumn{4}{c}{\textbf{AIR QUALITY}} & \multicolumn{4}{c}{\textbf{CONCRETE}} \\ 
\multicolumn{1}{l}{} & \multicolumn{4}{c}{Target Variable : NMHC (GT)} & \multicolumn{4}{c}{Target Variable : Compressive Strength} \\ 
\multicolumn{1}{l}{} & \multicolumn{4}{c}{Target Variable Range : [9.0 : 1189.0] } & \multicolumn{4}{c}{Target Variable Range : [2.33 : 82.6] } \\ 
\multicolumn{1}{l}{} & \multicolumn{4}{c}{ Prediction Count : 9,237 } & \multicolumn{4}{c}{Prediction Count : 910} \\ 
\cmidrule{3-5} \cmidrule{7-9} 
Metric &  & Exp. 3 - (a) & Exp. 3 - (b) & Exp. 3 - (c) &  & Exp. 4 - (a) & Exp. 4 - (b) & Exp. 4 - (c)  \\
\cmidrule{1-1} \cmidrule{3-5} \cmidrule{7-9} 
\begin{tabular}{lll}Drift  \\ Detector\\ \hspace{1mm} \end{tabular}  &  & \begin{tabular}{lll}RMSE \\ Abs. \\ Error\end{tabular} & \begin{tabular}{lll}ADWIN \& \\ RMSE Abs.\\ Error\end{tabular} & \begin{tabular}{lll}None \\ \hspace{1mm} \\ \hspace{1mm}\end{tabular} &  & \begin{tabular}{lll}RMSE \\ Abs. \\ Error\end{tabular} & \begin{tabular}{lll}ADWIN \& \\ RMSE Abs.\\ Error\end{tabular} & \begin{tabular}{lll}None \\ \hspace{1mm} \\ \hspace{1mm}\end{tabular} \\
\cdashline{3-5} \cdashline{7-9} \\
Model Updates &  & 20 & 1 & 0 &  & 3 & 0 & 0  \\
\cdashline{3-5} \cdashline{7-9}
\\
RMSE &  & \textbf{215.8276\textsuperscript{*}} & 217.8862 & 217.8862 &  & \textbf{17.2402\textsuperscript{*}} & 20.2035 & 20.2035  \\
\cdashline{3-5} \cdashline{7-9}
\\
Total Execution Time (Seconds) &  & 3.08 & 2.10 & 0.91 &  & 0.32 & 0.21 & 0.11  \\
\\
Processing Rate (Records/Second) &  & 2,999 & 4,398 & 10,150 &  & 2,843 & 4,333 & 8,272  \\
\midrule
\\

\multicolumn{1}{l}{} & \multicolumn{4}{c}{\textbf{PROTEIN}} & \multicolumn{4}{c}{\textbf{TURBINE}} \\ 
\multicolumn{1}{l}{} & \multicolumn{4}{c}{Target Variable : RMSD } & \multicolumn{4}{c}{Target Variable : TEY} \\ 
\multicolumn{1}{l}{} & \multicolumn{4}{c}{Target Variable Range : [0.0 : 20.999] } & \multicolumn{4}{c}{Target Variable Range : [100.02 : 179.5] } \\ 
\multicolumn{1}{l}{} & \multicolumn{4}{c}{ Prediction Count : 45,610 } & \multicolumn{4}{c}{Prediction Count : 36,613} \\ 
\cmidrule{3-5} \cmidrule{7-9} 
Metric &  & Exp. 5 - (a) & Exp. 5 - (b) & Exp. 5 - (c) &  & Exp. 6 - (a) & Exp. 6 - (b) & Exp. 6 - (c)  \\
\cmidrule{1-1} \cmidrule{3-5} \cmidrule{7-9} 
\begin{tabular}{lll}Drift  \\ Detector\\ \hspace{1mm} \end{tabular}  &  & \begin{tabular}{lll}RMSE \\ Abs. \\ Error\end{tabular} & \begin{tabular}{lll}ADWIN \& \\ RMSE Abs.\\ Error\end{tabular} & \begin{tabular}{lll}None \\ \hspace{1mm} \\ \hspace{1mm}\end{tabular} &  & \begin{tabular}{lll}RMSE \\ Abs. \\ Error\end{tabular} & \begin{tabular}{lll}ADWIN \& \\ RMSE Abs.\\ Error\end{tabular} & \begin{tabular}{lll}None \\ \hspace{1mm} \\ \hspace{1mm}\end{tabular} \\
\cdashline{3-5} \cdashline{7-9} \\
Model Updates &  & 3 & 0 & 0 &  & 38 & 12 & 0  \\
\cdashline{3-5} \cdashline{7-9}
\\
RMSE &  & \textbf{5.8719\textsuperscript{*}} & 6.0147 & 6.0147 &  & \textbf{8.6852\textsuperscript{*}} & 9.4496 & 9.4496  \\
\cdashline{3-5} \cdashline{7-9}
\\
Total Execution Time (Seconds) &  & 15.42 & 10.12 & 4.42 &  & 12.52 & 8.30 & 3.61  \\
\\
Processing Rate (Records/Second) &  & 2,957 & 4,506 & 10,319 &  & 2,924 & 4,411 & 10,142  \\
\midrule
\\

\multicolumn{1}{l}{} & \multicolumn{4}{c}{\textbf{TURBINE}} & \multicolumn{4}{c}{\textbf{TURBINE}} \\ 
\multicolumn{1}{l}{} & \multicolumn{4}{c}{Target Variable : CO } & \multicolumn{4}{c}{Target Variable : NOX} \\ 
\multicolumn{1}{l}{} & \multicolumn{4}{c}{Target Variable Range : [-1.048 : 18.443] } & \multicolumn{4}{c}{Target Variable Range : [-3.373 : 4.677] } \\ 
\multicolumn{1}{l}{} & \multicolumn{4}{c}{ Prediction Count : 36,613 } & \multicolumn{4}{c}{Prediction Count : 36,613} \\ 
\cmidrule{3-5} \cmidrule{7-9} 
Metric &  & Exp. 7 - (a) & Exp. 7 - (b) & Exp. 7 - (c) &  & Exp. 8 - (a) & Exp. 8 - (b) & Exp. 8 - (c)  \\
\cmidrule{1-1} \cmidrule{3-5} \cmidrule{7-9} 
\begin{tabular}{lll}Drift  \\ Detector\\ \hspace{1mm} \end{tabular}  &  & \begin{tabular}{lll}RMSE \\ Abs. \\ Error\end{tabular} & \begin{tabular}{lll}ADWIN \& \\ RMSE Abs.\\ Error\end{tabular} & \begin{tabular}{lll}None \\ \hspace{1mm} \\ \hspace{1mm}\end{tabular} &  & \begin{tabular}{lll}RMSE \\ Abs. \\ Error\end{tabular} & \begin{tabular}{lll}ADWIN \& \\ RMSE Abs.\\ Error\end{tabular} & \begin{tabular}{lll}None \\ \hspace{1mm} \\ \hspace{1mm}\end{tabular} \\
\cdashline{3-5} \cdashline{7-9} \\
Model Updates &  & 1,162 & 15 & 0 &  & 1,197 & 15 & 0  \\
\cdashline{3-5} \cdashline{7-9}
\\
RMSE &  & 1.2819 & \textbf{1.2344\textsuperscript{*}} & \textbf{1.2344\textsuperscript{*}} &  & \textbf{1.8815\textsuperscript{*}} & 2.0484 & 2.0484  \\
\cdashline{3-5} \cdashline{7-9}
\\
Total Execution Time (Seconds) &  & 12.68 & 8.20 & 3.56 &  & 12.52 & 8.27 & 3.60 \\
\\
Processing Rate (Records/Second) &  & 2,887 & 4,465 & 10,284 &  & 2,924 & 4,427 & 10,170  \\
\bottomrule
\bottomrule
\end{tabular}
\end{adjustbox}
\vspace{1mm}
\caption{Linear Regression Results of the 24 Experiments - (Standardized Input -  Best Result\textsuperscript{*}).}
\label{tab:lr_results}
\end{table}

\subsection{Overall observations}
In summary, the \textit{RMSE Absolute Error} framework method was the best at predicting target variables for all experiments with the exception of experiment 7-(a). In predicting Turbine CO values, the other methods produced better results. Generally, the RMSE performance metric improved when a model was retrained. In all cases, the preferable predictive performance of the best model came at a cost in terms of added execution time. Retraining the a model after a drift has been detected adds to the overall execution time. Balancing the precision of predictions and the processing time in stream processing involves a trade-off, but within the scope of our experiments, it represents a cost-effective compromise. This holds particularly true considering that the longest execution time for any experiment slightly exceeds 15 seconds, and this experiment involved the largest dataset in our study.


\subsection{Limitations}
\label{sec:limitations}
In our experiments, we employed a local, simulated data stream, which essentially entails iteratively traversing a data file and sequentially providing input values for model training and inference. This process eliminated the, often real-world, need of transmitting data to and from the inference process and contributed to an overall reduction in the execution time of every experiment by avoiding any associated network traffic overhead. In a real-world cloud deployment, there would be network traffic overhead. Even in an edge computing context, despite the benefits of localized data processing in minimizing data transfers, there remains certain network overhead and computation on a potentially resource-constrained device. These factors could contribute to overall increases in execution times that are not captured in our experiments.

The threshold value is currently not adaptive. As described in Section~\ref{sec:adaptive_regresion_model}, after the initial training phase, our framework uses a sliding window and buffer to process new, unlabeled data as it arrives. One subset of the sliding window is used for model fitting. The other subset (i.e. the buffer) is used for testing and calculating intermediary RMSE values, which are used to determine drifts in the data. With this drift detection approach, new intermediary RMSE values are compared with the most recently stored intermediary RMSE value using a pre-determined threshold. If a drift is detected (i.e. the RMSE difference is larger than the threshold), the regression model is retrained. The threshold parameter is fixed and needs to be determined in advance. This limitation can be mitigated by enhancing the drift determination approach with multiple sliding windows in conjunction with standard deviation values to update the threshold dynamically. This approach will be explored in future work.

The RMSE evaluation metric improvements that our proposed framework affords are small in some experiments. In others, they are more significant. It's crucial to evaluate predictive performance improvements in practical terms and to consider the context in which model predictive performance improvements occur. For example, in fields such as finance or healthcare, where predictions have significant consequences, any improvement in accuracy, no matter how small, can be valuable. Moreover, enhancements in predictive accuracy may necessitate a rise in computational resources, and the trade-off between these improvements and the associated costs must be carefully considered. In this study, the adoption of an online adaptive learning approach using a simple regression model resulted in a need for only modest computing capacity. This approach facilitated a more straightforward model deployment and update context when compared to alternatives like the batch context.

\section{Conclusions and Future Work}
\label{sec:conclusion}
In this study, we present a significant advancement in the realm of unsupervised regression for streaming data. Our contribution lies in the introduction of a simple, explainable, and adaptive regression model tailored specifically for dynamic data streams. The proposed model is accompanied by a pioneering drift detection method, collectively addressing critical challenges and laying the groundwork for transformative applications in real-world scenarios. In order to assess the effectiveness of our approach, we conduct comparative analysis of the proposed method relative to a non-adapting baseline.

The positive outcomes derived from our comprehensive evaluation affirm the practical impact and effectiveness of our online, unsupervised regression framework. Noteworthy improvements over the non-adapting baseline  underscore the robustness of our approach and its potential to elevate predictive performance in various domains.

The implications of our framework extend prominently to sectors with high-stakes predictions, such as finance and healthcare. Even incremental enhancements in accuracy, as demonstrated by our framework, carry substantial value in contexts where predictions wield significant consequences. This underlines the relevance and importance of our framework's advancements, particularly in domains grappling with limited labeled data. 

Looking forward, our future work will venture into dynamic thresholding—an innovation poised to imbue our framework with real-time adaptability to changes in underlying data patterns. This dynamic approach, guided by multiple intermediary RMSE values, promises enhanced adaptability, positioning our framework as a flexible and responsive tool for evolving conditions. Future work will compare alternative streaming algorithms that also consider concept drift. Finally, the evaluation will be extended to many include additional datasets.

While we acknowledge the inherent limitations discussed in Section~\ref{sec:limitations}, our study stands as a catalyst for further exploration in the field. Researchers are encouraged to delve into unexplored facets, with a particular focus on the intriguing realm of human-in-the-loop model updating. This invites a collaborative effort to refine and expand upon our initial approach, fostering continual progress in the evolving landscape of adaptive unsupervised regression.

\bibliographystyle{IEEEtran}
\bibliography{IEEEabrv, bibliography}

\newpage

\appendix
\section{Appendix}

\subsection{Air Quality - (Exp. 1) -  Target Variable :  CO (GT)}
\label{ap:1}
\FloatBarrier
\begin{figure}[!ht] 
  \centering
  \subfloat[RMSE Abs. Error Drift Detector \label{fig:aq_cogt_pred_true_rmse}]{
       \includegraphics[height=4.6cm]{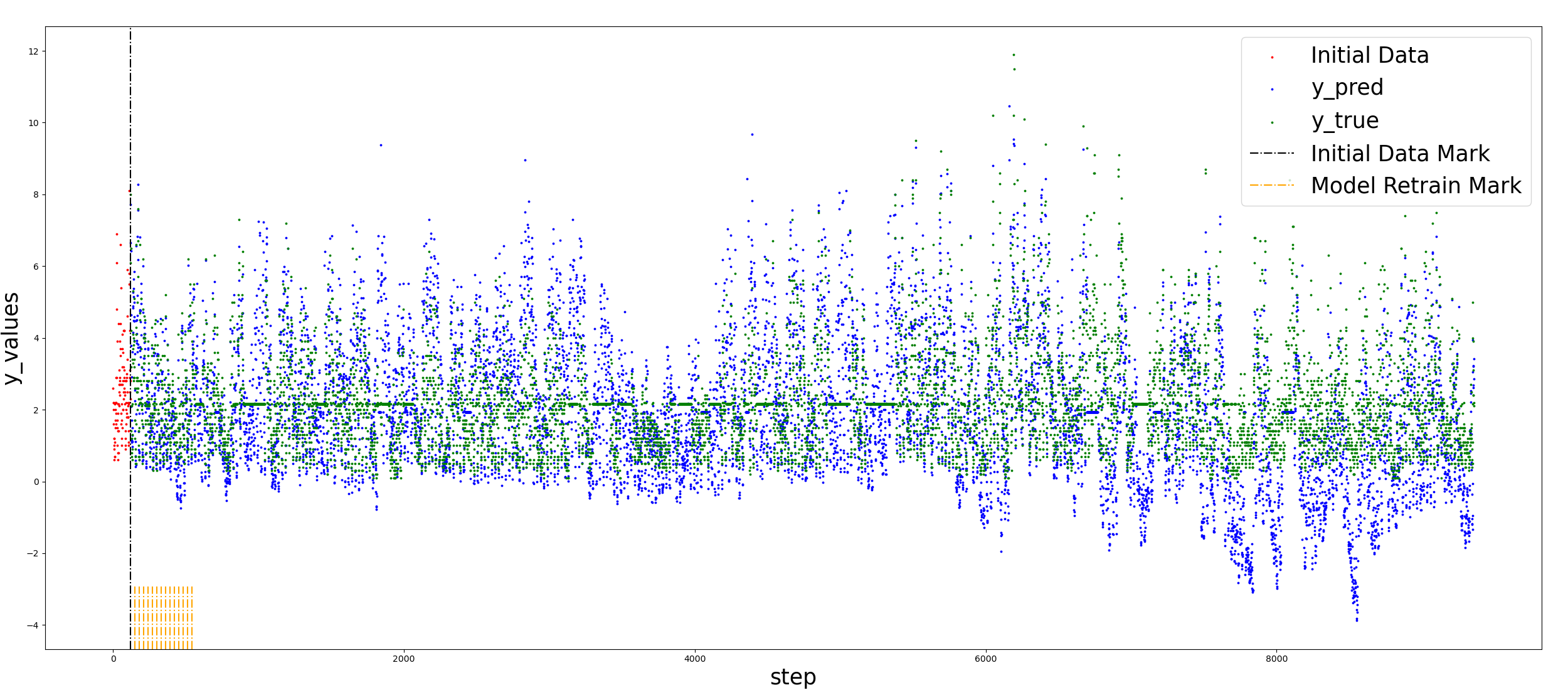}}
  \hfill
  \subfloat[ADWIN \& RMSE Abs. Error Drift Detector \label{fig:aq_cogt_pred_true_adwin_rmse}]{
       \includegraphics[height=4.6cm]{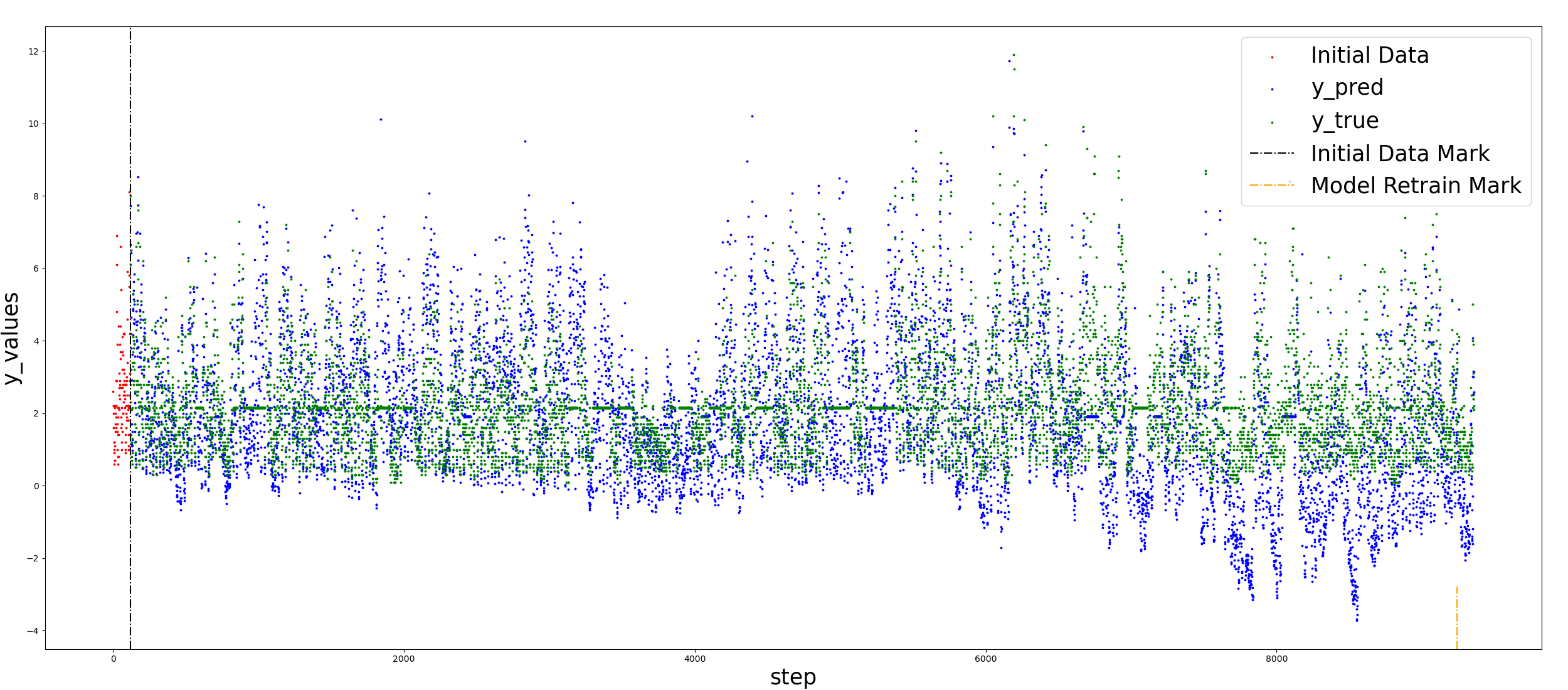}}
   \hfill
   \subfloat[No Drift Detector \label{fig:aq_cogt_pred_true_noe}]{
       \includegraphics[height=4.6cm]{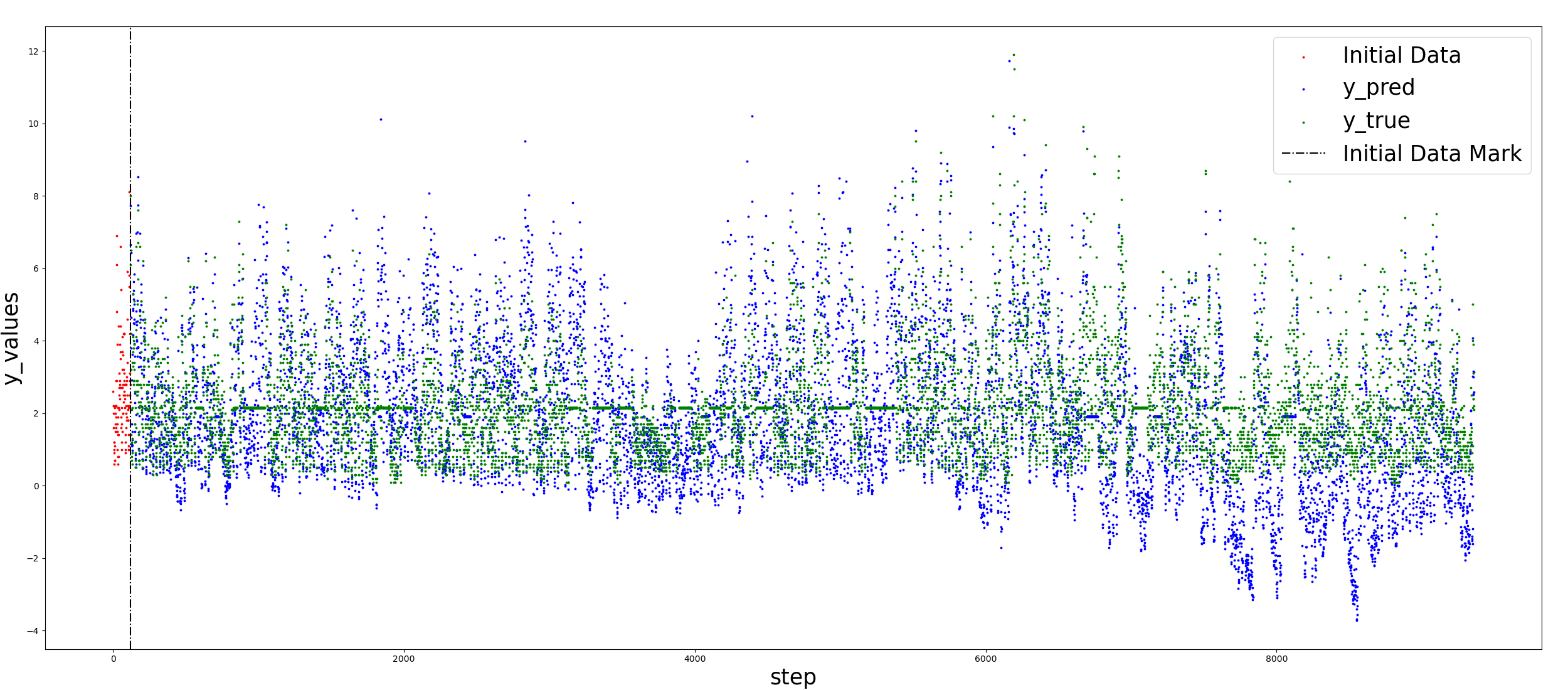}}
  \label{fig:aq_cogt_results} 
  \caption{Air Quality CO (GT) - Predicted and Ground Truth Values}
\end{figure}
\FloatBarrier

\newpage

\subsection{Concrete - (Exp. 4) -  Target Variable :  Compressive Strength}
\label{ap:2}
\FloatBarrier
\begin{figure}[!ht] 
  \centering
  \subfloat[RMSE Abs. Error Drift Detector \label{fig:concrete_pred_true_rmse}]{
       \includegraphics[height=4.9cm]{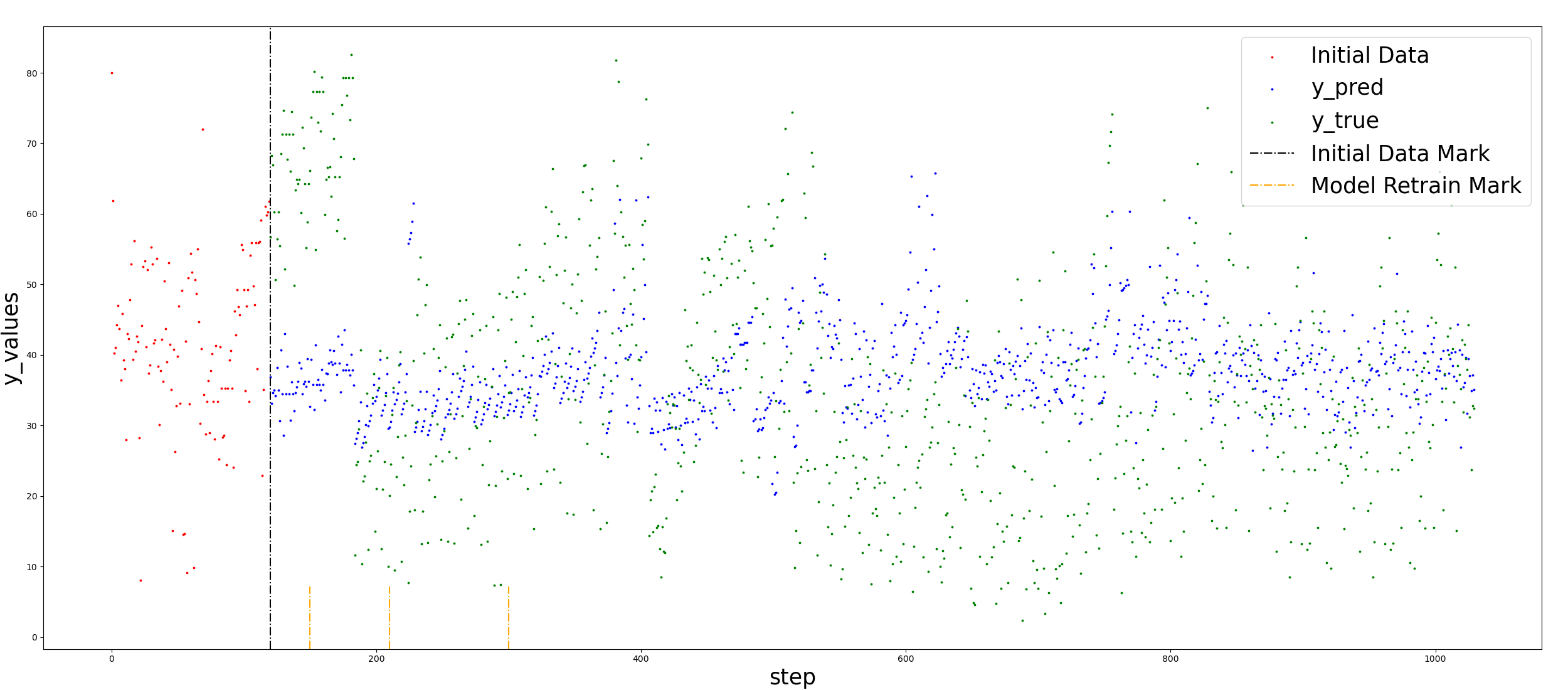}}
  \hfill
  \subfloat[ADWIN \& RMSE Abs. Error Drift Detector \label{fig:concrete_pred_true_adwin_rmse}]{
       \includegraphics[height=4.9cm]{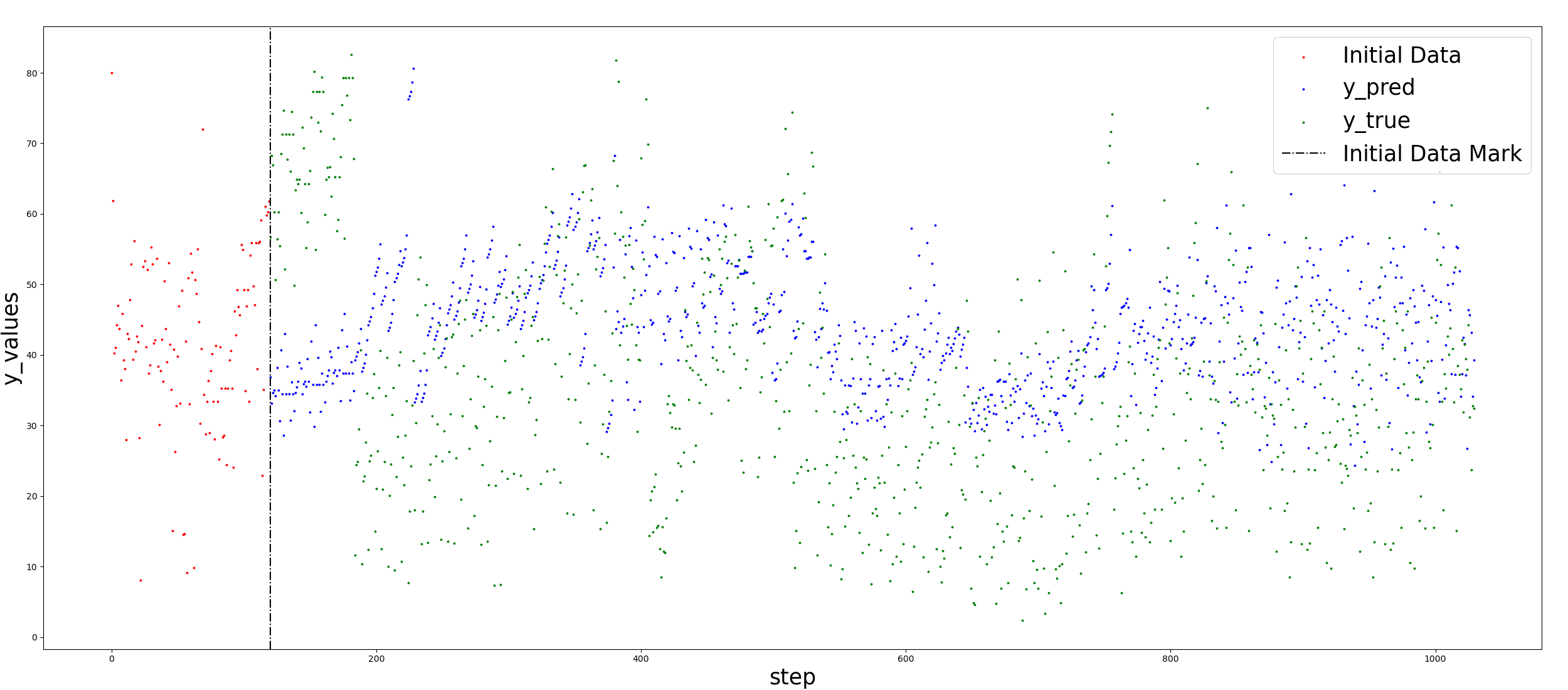}}
   \hfill
   \subfloat[No Drift Detector \label{fig:concrete_pred_true_noe}]{
       \includegraphics[height=4.9cm]{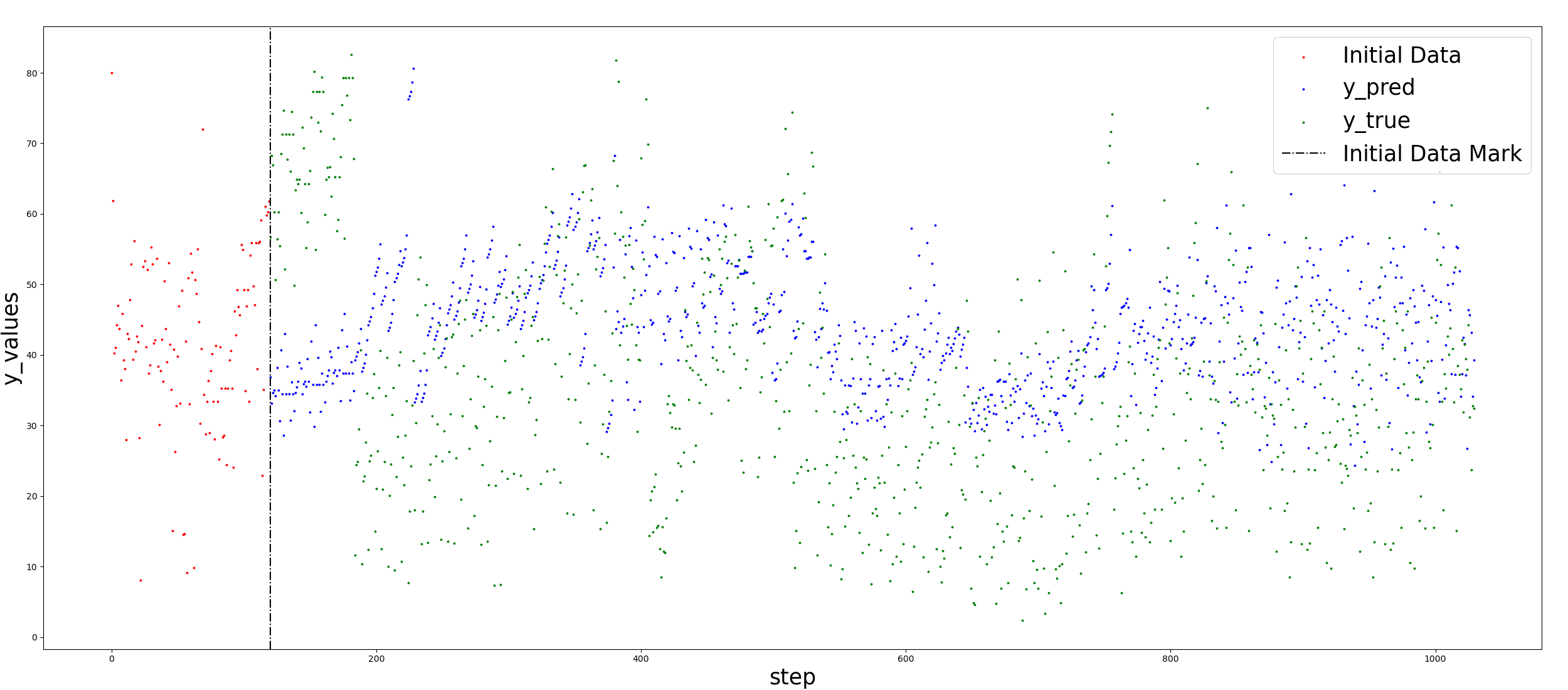}}
  \label{fig:concrete_results} 
  \caption{Concrete Compressive Strength - Predicted and Ground Truth Values}
\end{figure}
\FloatBarrier

\newpage

\subsection{Protein - (Exp. 5) -  Target Variable :  RMSD}
\label{ap:3}
\FloatBarrier
\begin{figure}[!ht] 
  \centering
  \subfloat[RMSE Abs. Error Drift Detector \label{fig:protein_pred_true_rmse}]{
       \includegraphics[height=4.9cm]{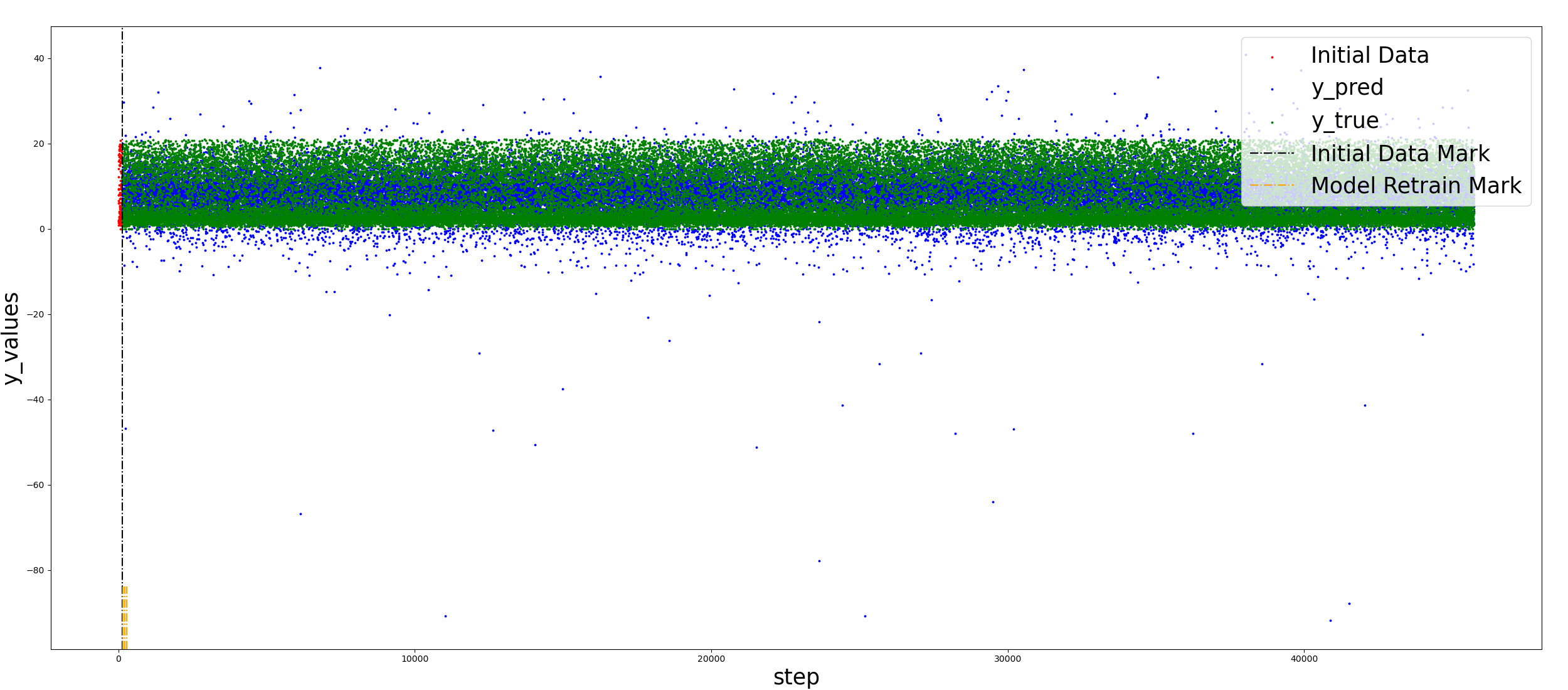}}
  \hfill
  \subfloat[ADWIN \& RMSE Abs. Error Drift Detector \label{fig:protein_pred_true_adwin_rmse}]{
       \includegraphics[height=4.9cm]{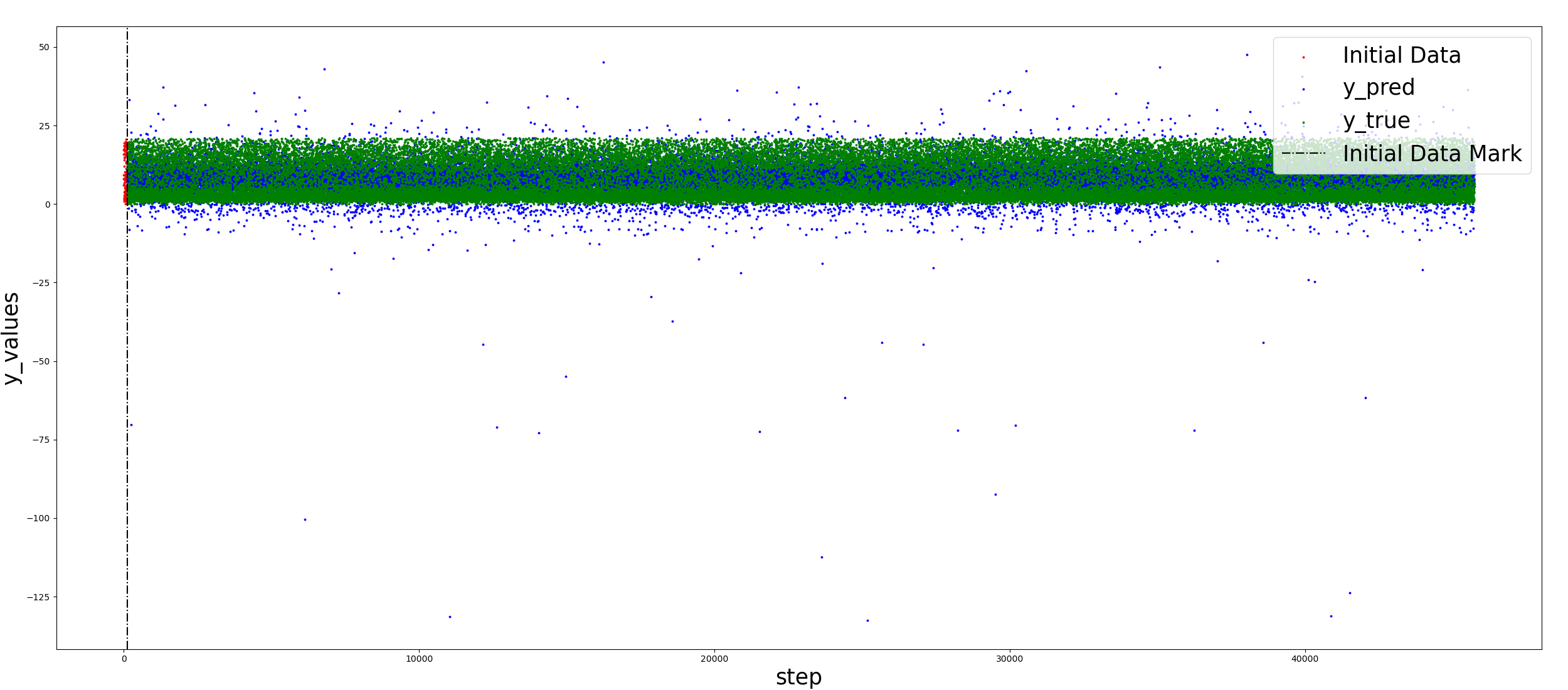}}
   \hfill
   \subfloat[No Drift Detector \label{fig:protein_pred_true_noe}]{
       \includegraphics[height=4.9cm]{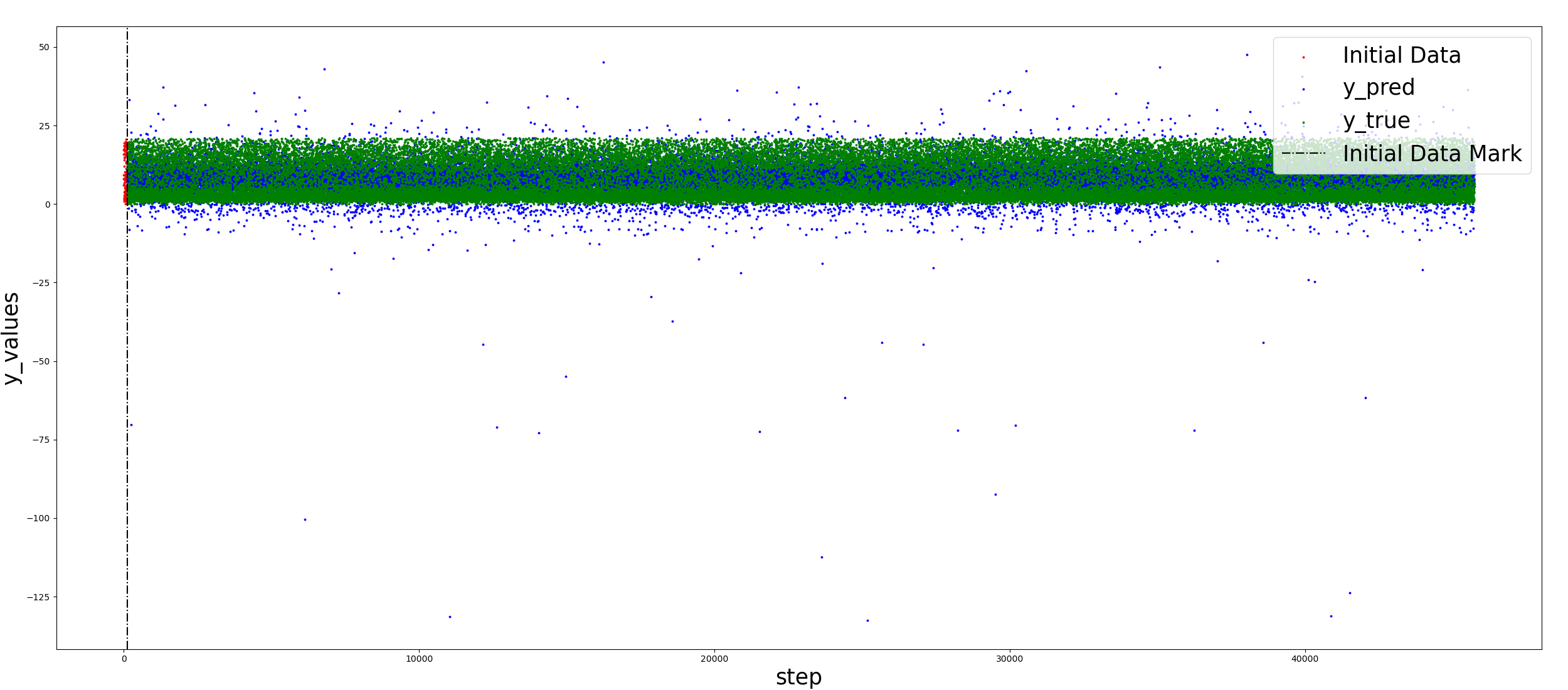}}
  \label{fig:protein_results} 
  \caption{Protein RMSD - Predicted and Ground Truth Values}
\end{figure}
\FloatBarrier

\newpage

\subsection{Turbine - (Exp. 6) -  Target Variable :  TEY}
\label{ap:4}
\FloatBarrier
\begin{figure}[!ht] 
  \centering
  \subfloat[RMSE Abs. Error Drift Detector \label{fig:turb_tey_pred_true_rmse}]{
       \includegraphics[height=4.9cm]{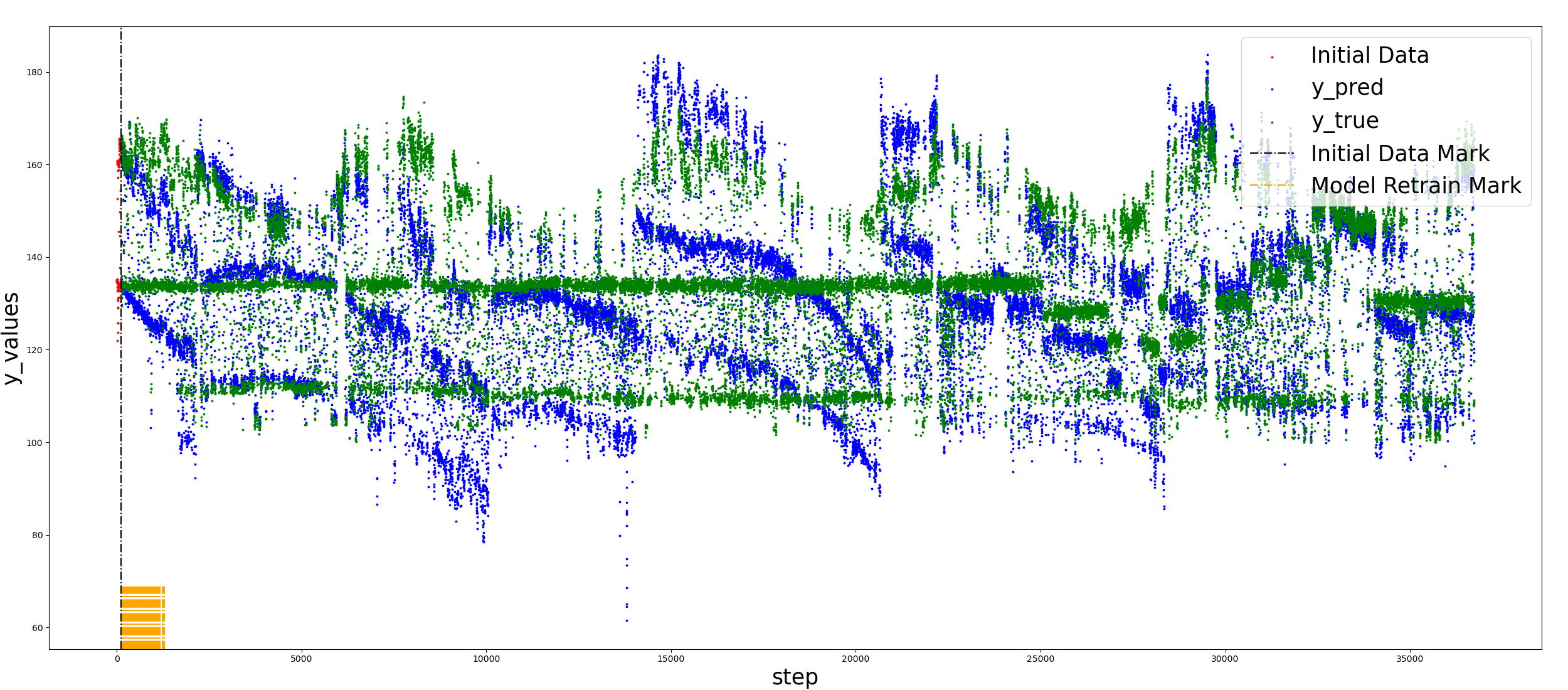}}
  \hfill
  \subfloat[ADWIN \& RMSE Abs. Error Drift Detector \label{fig:turb_tey_pred_true_adwin_rmse}]{
       \includegraphics[height=4.9cm]{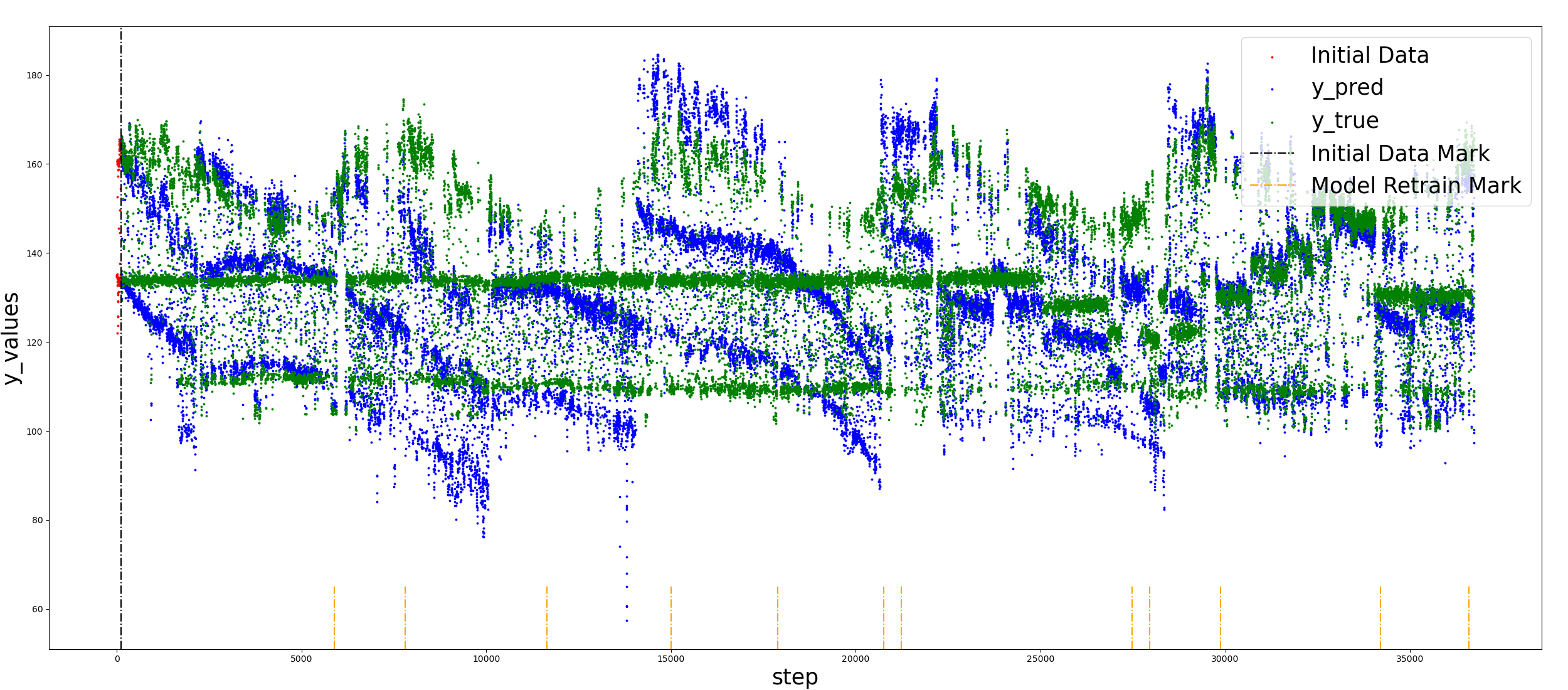}}
   \hfill
   \subfloat[No Drift Detector \label{fig:turb_tey_pred_true_noe}]{
       \includegraphics[height=4.9cm]{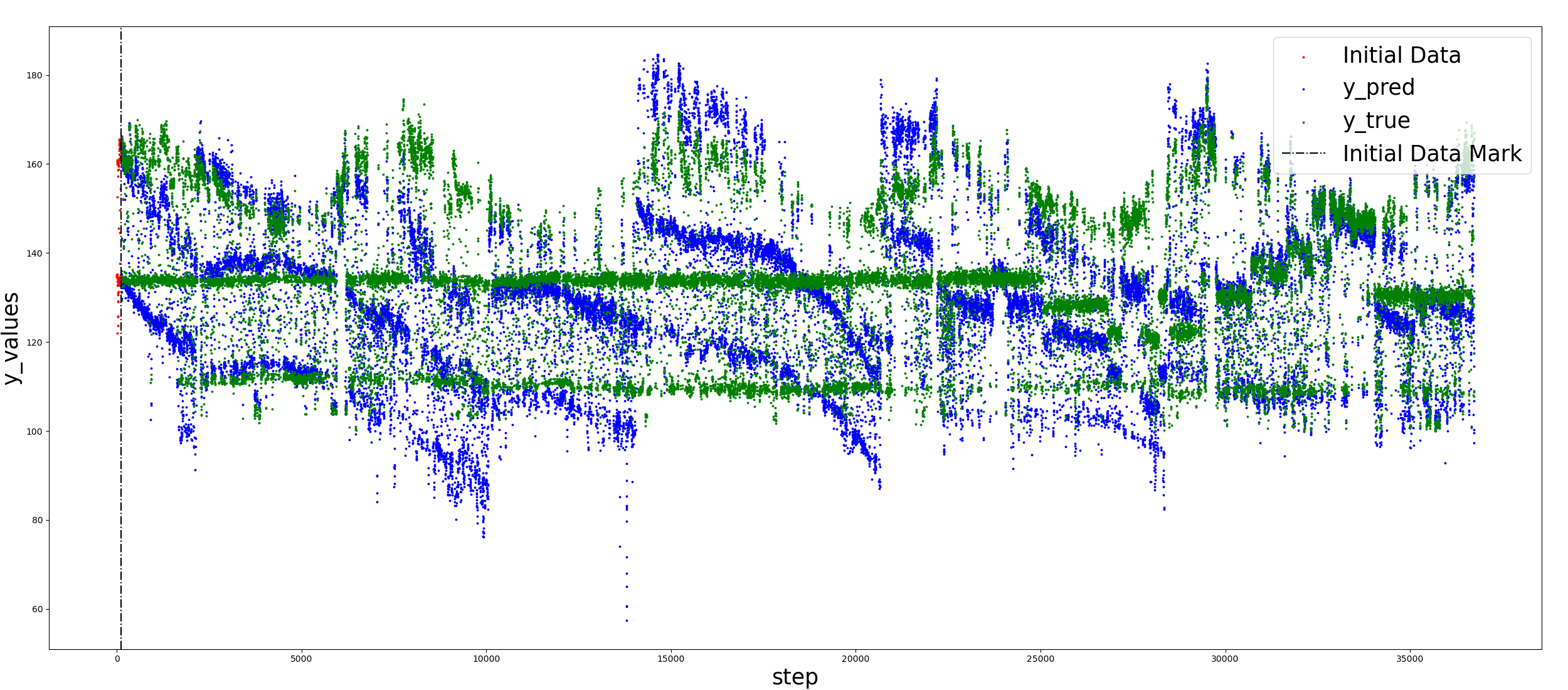}}
  \label{fig:turb_tey_results} 
  \caption{Turbine TEY - Predicted and Ground Truth Values}
\end{figure}
\FloatBarrier

\end{document}